\documentclass[]{article}
\usepackage{lmodern}
\usepackage{amssymb,amsmath}
\usepackage{ifxetex,ifluatex}
\usepackage{fixltx2e} 
\ifnum 0\ifxetex 1\fi\ifluatex 1\fi=0 
  \usepackage[T1]{fontenc}
  \usepackage[utf8]{inputenc}
\else 
  \ifxetex
    \usepackage{mathspec}
  \else
    \usepackage{fontspec}
  \fi
  \defaultfontfeatures{Ligatures=TeX,Scale=MatchLowercase}
\fi
\IfFileExists{upquote.sty}{\usepackage{upquote}}{}
\IfFileExists{microtype.sty}{%
\usepackage{microtype}
\UseMicrotypeSet[protrusion]{basicmath} 
}{}
\usepackage[margin=1in]{geometry}
\usepackage{hyperref}
\hypersetup{unicode=true,
            pdftitle={RedDwarfData: a simplified dataset of StarCraft matches},
            pdfborder={0 0 0},
            breaklinks=true}
\usepackage{graphicx,grffile}
\makeatletter
\def\maxwidth{\ifdim\Gin@nat@width>\linewidth\linewidth\else\Gin@nat@width\fi}
\def\maxheight{\ifdim\Gin@nat@height>\textheight\textheight\else\Gin@nat@height\fi}
\makeatother
\setkeys{Gin}{width=\maxwidth,height=\maxheight,keepaspectratio}
\IfFileExists{parskip.sty}{%
\usepackage{parskip}
}{
\setlength{\parindent}{0pt}
\setlength{\parskip}{6pt plus 2pt minus 1pt}
}
\ifx\paragraph\undefined\else
\let\oldparagraph\paragraph
\renewcommand{\paragraph}[1]{\oldparagraph{#1}\mbox{}}
\fi
\ifx\subparagraph\undefined\else
\let\oldsubparagraph\subparagraph
\renewcommand{\subparagraph}[1]{\oldsubparagraph{#1}\mbox{}}
\fi

\let\rmarkdownfootnote\footnote%
\def\footnote{\protect\rmarkdownfootnote}

\usepackage{titling}

  \title{RedDwarfData: a simplified dataset of StarCraft matches}
  \pretitle{\vspace{\droptitle}\centering\huge}
  \posttitle{\par}
  \author{Juan J. Merelo-Guervós, Antonio Fernández-Ares\\
University of Granada, Spain\\[2\baselineskip]Antonio Álvarez
Caballero\\
University of Oviedo, Spain\\[2\baselineskip]Pablo García-Sánchez\\
University of Cádiz, Spain\\[2\baselineskip]Victor Rivas\\
University of Jaen, Spain}
  \preauthor{\centering\large\emph}
  \postauthor{\par}
  \predate{\centering\large\emph}
  \postdate{\par}
  \date{December 29th, 2017}

\begin{document}
\maketitle
\begin{abstract}
The game Starcraft is one of the most interesting arenas to test new
machine learning and computational intelligence techniques; however,
StarCraft matches take a long time and creating a good dataset for
training can be hard. Besides, analyzing match logs to extract the main
characteristics can also be done in many different ways to the point
that extracting and processing data itself can take an inordinate amount
of time and of course, depending on what you choose, can bias learning
algorithms. In this paper we present a simplified dataset extracted from
the set of matches published by Robinson and Watson, which we have
called RedDwarfData, containing several thousand matches processed to
frames, so that temporal studies can also be undertaken. This dataset is
available from GitHub under a free license. An initial analysis and
appraisal of these matches is also made.
\end{abstract}

\section{Introduction}\label{introduction}

\emph{StarCraft} is a well-known Real Time Strategy (\emph{RTS}) game
whose first version was released in 1998 (Yoon 2001). The game consists
of two opponents trying to beat each other over a square-shaped arena.
Every player has to spawn workers and create buildings using the
available resources, which are located over the terrain. Players can use
one of three different kinds of species, called Zergs, Protoss or
Terrans, with different kinds of individuals, which differ in their
skills, so that players have a wide range of possibilities in order to
plan their strategies. The need to dynamically take decisions for both
the short and the long terms make RTS adequate scenarios to test AI
algorithms (Ontanón et al. 2013) that could be used later in real
problems including financial, military and logistics scenarios.

\emph{StarCraft} has become popular among researchers thanks in part to
\emph{BWAPI}, an API developed by the community that allows programming
artificial agents and that was developed for StarCraft competitions
(Buro and Churchill 2012). These agents can act just like regular
players, according to a previously established algorithm; but, more
importantly, they can collect data as matches are running. The data
collected this way can be later analysed, so that machine learning
techniques can be applied for both supervised and unsupervised learning.

Developing and releasing datasets for developing and applying these
techniques is thus a way of expanding research in this area and allowing
researchers to develop their own bots or prediction algorithms without
having to scrape them from websites or actually play the games, a
time-consuming task, to extract data from them. Several researchers have
done so and we will refer to them in the state of the art; in this paper
we start from Robertson and Watson's dataset (Robertson and Watson 2014)
and create a simpler one that can be used without needing many
resources; even so, the data set is balanced with respect to duration,
races and number of resources used by the players.

The rest of the paper is organized as follows: next we will present the
Starcraft datasets that have been published so far. In the next section
we will show how Robertson and Watson's dataset was processed and the
rationale for doing it that way. Then we will present a macro picture of
the data set, to eventually draw some conclusions from it and possible
lines of work using this dataset.

\subsection{State of the art}\label{state-of-the-art}

The main motivation for writing this paper was the proliferation of new
StarCraft datasets in the last few years. Players have been uploading
their matches command logs to several websites, such as TeamLiquid
\footnote{\url{http://www.teamliquid.net}} or BWReplays
\footnote{\url{http://www.bwreplays.com}} for over 10 years. These logs
are usually employed by the player community to learn new playing
techniques from other players. From the point of view of using them for
AI training, however, these datasets are in raw form and require some
kind of command execution, parsing, data selection and filtering in
order to serve as a good base for learning models (Lin et al. 2017).
Moreover, a lot of uploaded replays may be corrupt or incomplete.
Different authors have been processing these replays to create datasets
that can be used to predict the outcome of a match or a battle or simply
gather insights on gaming strategies. The first paper that describes the
processing of replays was the one presented by Weber and Mateas (Weber
and Mateas 2009), where they processed replays to create a dataset to
compare several classification algorithms. 5393 replays were used and
they store specific events during the games, not the full game state,
validating the possibility of predicting the winner of a game to a
certain degree. Later, other datasets that store the full game state to
apply generic machine learning techniques have appeared (Synnaeve and
Bessière 2012).

Instead providing the game logs directly, Robertson and Watson
(Robertson and Watson 2014) distributed the logs in a database, created
by obtaining the replays from several replays webpages, executing the
game and storing a large number of game features data every certain
number of frames. This dataset has been used by other researchers, for
example, to create models differentiating by race (Ravari, Bakkes, and
Spronck 2016).

Recently, and built upon the Synnaeve and Bessiere work (Synnaeve and
Bessière 2012), Lin et al. (Lin et al. 2017) have released the largest
dataset to date. It includes 65646 replays, storing the full state of
the game every 3 frames, including up to 30 features by unit and battle
detections, storing all the data in TorchCraft format to be accessed by
specific software clients. Even using compression mechanisms, this
dataset requires 365GB of space. The aim of using this dataset is to
apply Deep Learning algorithms that require a large amount of data.

However, there is not a small and limited dataset ready to use for
teaching purposes, for example, to directly use as input for certain
algorithms without any modification. This is the reason we have used the
Robertson and Watson database (Robertson and Watson 2014) to extract a
limited number of features and parsed it directly to CSV (Comma
Separated Values) format. These CSV files can be easily downloaded and
straightforwardly parsed with any programming language.

\subsection{Data extraction and
filtering}\label{data-extraction-and-filtering}

One of the advantages of the presented dataset is that they are already
pre-processed and ready to use. The origin of the samples are the
replays obtained by (Robertson and Watson 2014). As previously stated,
they provided the data in the form of a huge relational database, so an
intermediate step to produce an usable and manageable CSV file was
needed.

For every available match in the database, identified by it
\emph{ReplayID}, a different number of samples is extracted. The first
step is to take the \emph{Duration} and the identifiers
\emph{PlayerReplayID} of the two players involved in the game. The
\emph{PlayerReplayID} with the minimum value is taken as the first
player; this does not actually introduce a bias.

Because of the dynamic nature of the game, it is important to use a time
reference to be able to track the match in every moment. Instead
sampling the matches every certain number of frames as other authors
propose, the chosen reference is the \emph{Frame} when a change in
resources (\emph{Minerals}, \emph{Gas} and \emph{Supply}) occurs. This
instant does not have to be the same for the two players, so all the
instants when a change is discovered are saved. This choice implies a
lot of empty values in the database; however, this is easily fixed
taking the last known value to fill each empty one: if there is not a
change for a player, the resources would be the last seen ones.

This decision is also important to track the units, that are valued by
their price in resources. However, the value recorded in the original
database is split by regions, which are pieces of the map. To register
the value of some units, all the region's values have to be summed up,
but the frame when these changes occur are not the same that the
resources' ones, so the last change for each region is saved and
accumulated. The same process is also performed for the enemy units.
However, this value is an estimation, not a real one, because due to the
rules, a player do not know nothing about the enemy unless they see the
other, and at that moment, these real values are stored in the logs.

With all the data well-formed, only the \emph{ReplayID}, \emph{Winner}
and \emph{Duration} have to be added to the match. The data of a game
does not have any missing values, so they are ready to be saved. As it
was said before, this process has to be repeated for each match.

\subsection{The big picture}\label{the-big-picture}

A few macro analysis of the data in this dataset will allow us to
compare it with other datasets and also appraise its balance and general
features. The data set, which is available
\href{https://github.com/analca3/StarCraft-winner-prediction/tree/master/data}{from
GitHub under a free license}, includes 4655 different matches with
matches among all different races available in Starcraft.

The duration of the matches is represented below.

\includegraphics{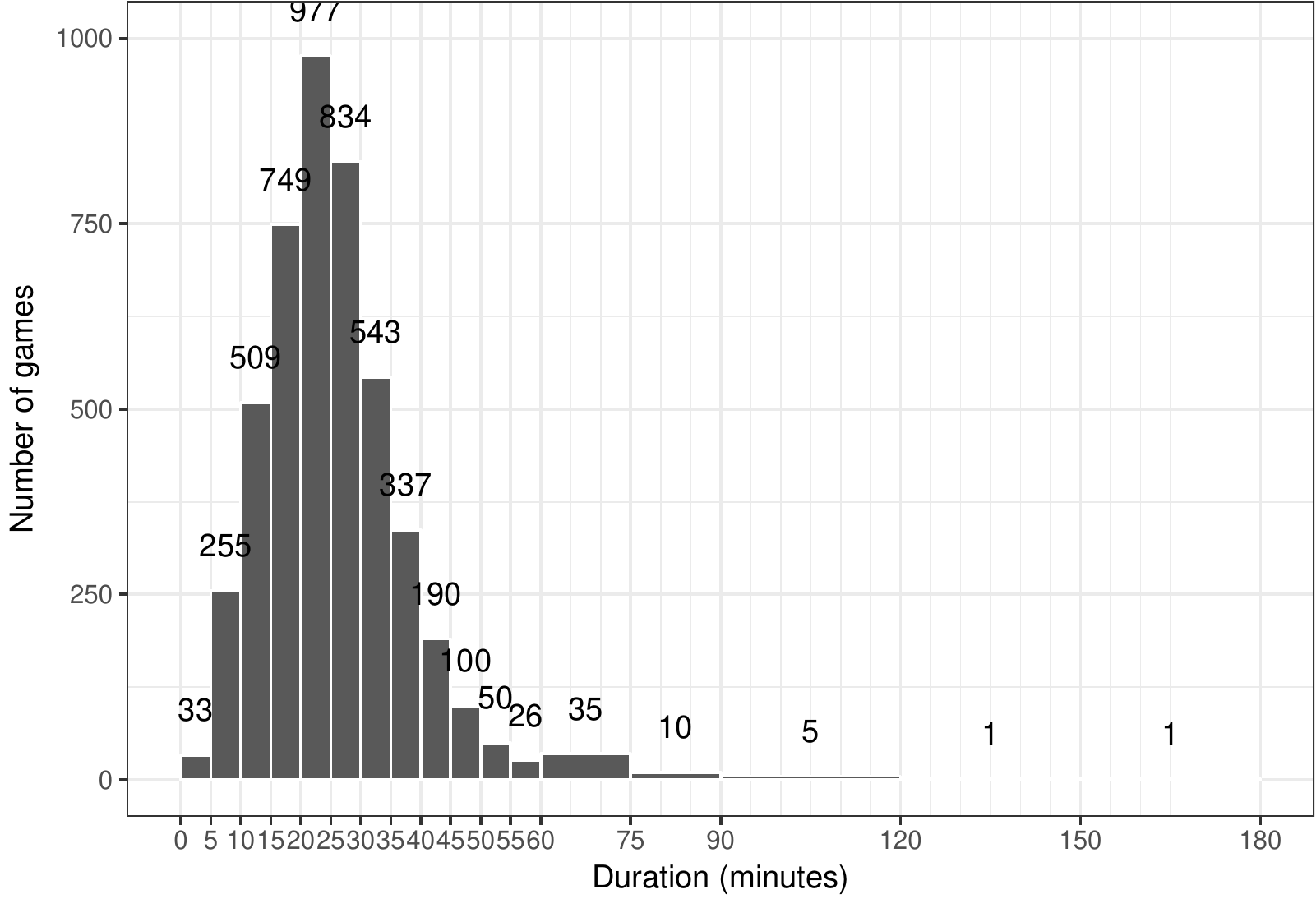}

The mode is between 20 and 25 minutes, with average equal to 25.2570211
and median 25.2570211; they are not too different which indicates that
the distribution is not too skewed.

\includegraphics{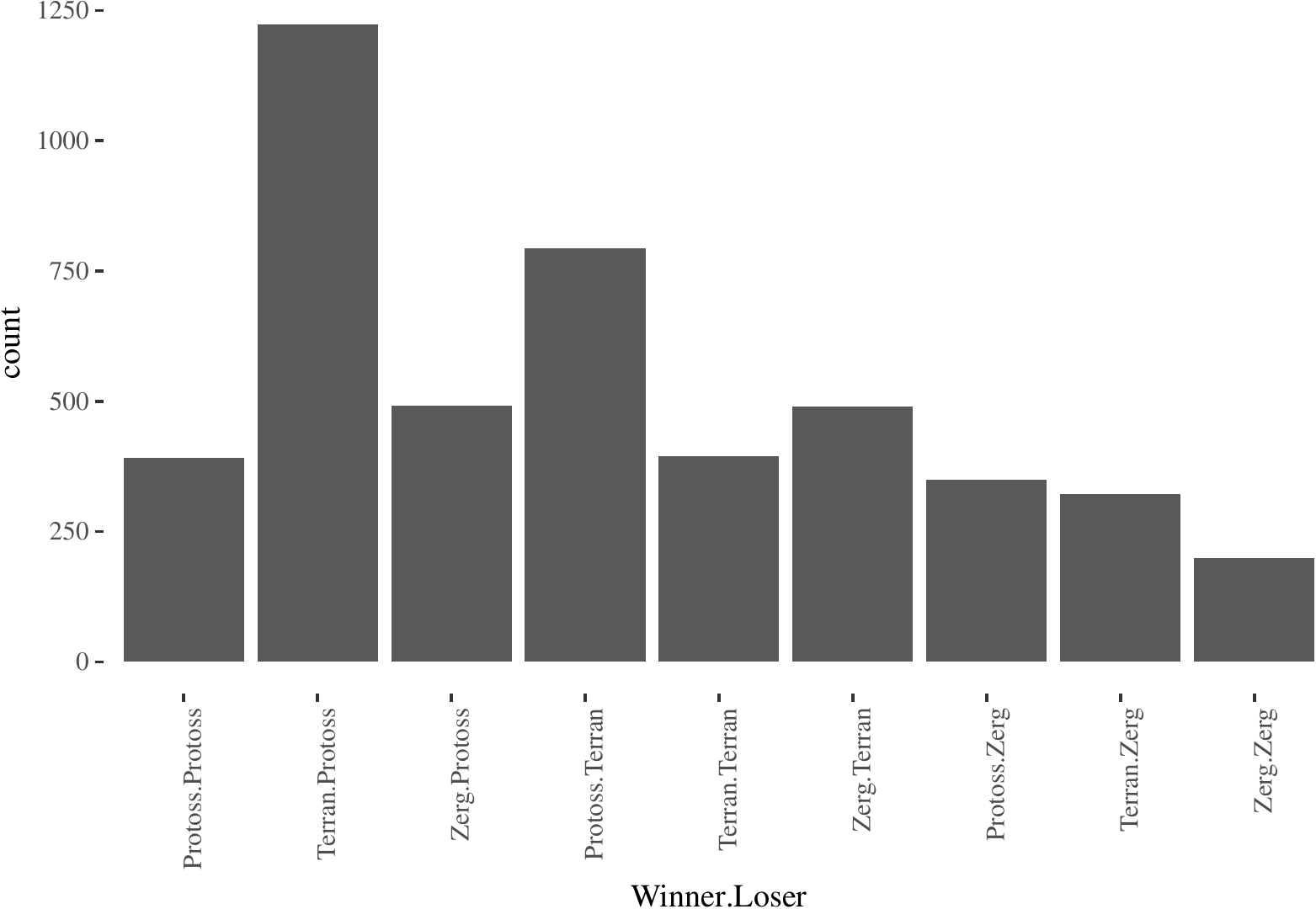} The
combination Terran vs.~Protoss includes the bigger number of matches,
with close to two thousands, of which the majority are won by the
Terran. But in fact, the race that wins the most games it is involved it
are the Zerg, as is shown below.

\includegraphics{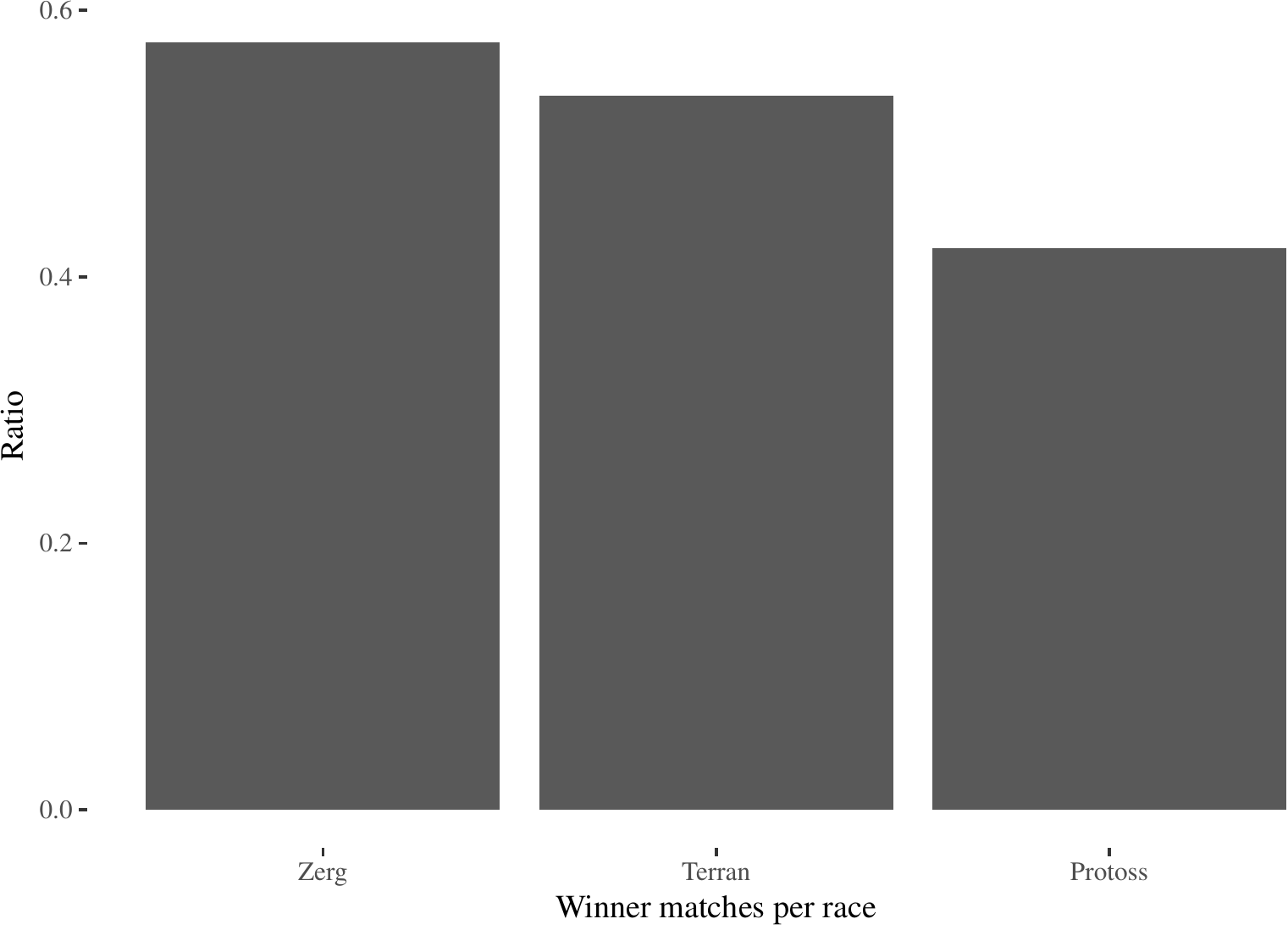}

Both macro measures show that the data set is well balanced, and
includes battles in all possible combinations. On the other hand, we can
use match data to find out if the duration of the match is affected by
the combination of races

\includegraphics{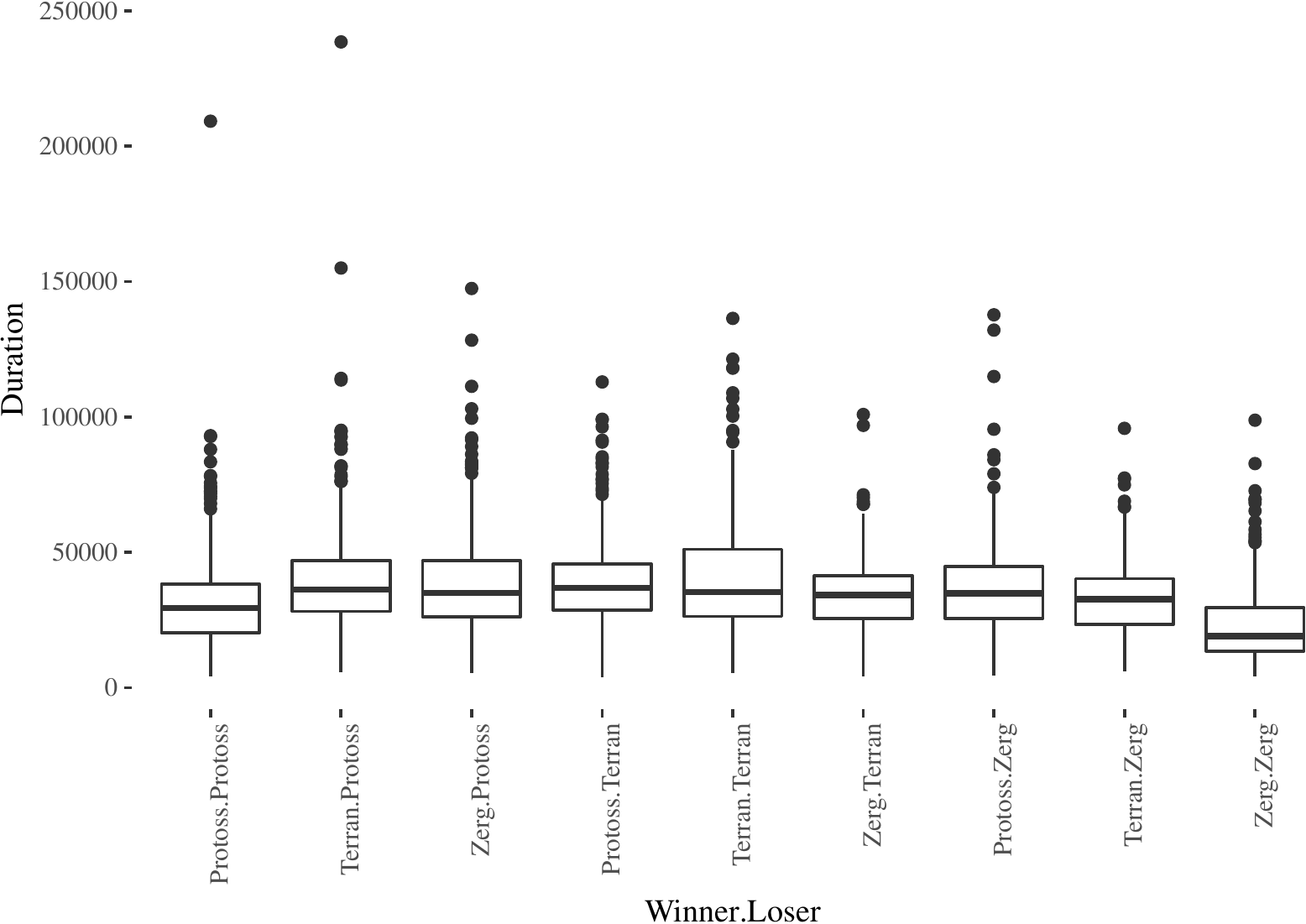}

And the chart shows that, except for the matches that include Zerg
against Zerg and Protoss vs.~Protoss, duration is quite similar and is
more affected by the game dynamics that by differences or similarities
between races.

\subsection{Conclusions}\label{conclusions}

Our intention with the release of RedDwarfData and this report was to
introduce a new lightweight dataset that includes matches by all races
and that can be used for drawing conclusions about Starcraft matches, as
well as use them to actually train bots that play Starcraft. An initial
exploratory analysis draws the conclusion that the Zerg might be the
best race to use in a game bot; however, this macro conclusion will have
to be qualified by an actual analysis of the strategies generally used
by Zergs in the particular games included in the data set. RedDwarfData
and the scripts that are used to process it are hosted
\href{https://github.com/analca3/StarCraft-winner-prediction}{in Github:
\texttt{https://github.com/analca3/StarCraft-winner-prediction}}. This
paper is hosted also in GitHub, and includes the scripts used to
generate aggregated data and the data that has been used to generate
this paper, which is written in R Markdown and includes the scripts to
generate these charts. It has a Apache license and can be downloaded
\href{https://github.com/geneura-papers/2017-StarCraft-Data}{from GitHub
at \texttt{https://github.com/geneura-papers/2017-StarCraft-Data}}. This
dataset has been used already in (Álvarez-Caballero et al. 2017) to make
early predictions of the outcome of the match, in order to shorten the
time needed to evaluate bots in an optimization environment. Our
intention is to create surrogate models of matches so that we can use
evolutionary algorithms to create game bots that can be competitive in
this game.

\subsection{Acknowledgements}\label{acknowledgements}

This paper has been funded in part by the Spanish national excellence
project TIN2014-56494-C4-3-P (EphemeCH). We acknowledge also the support
of TIN2017-85727-C4-2-P (DeepBio).

\subsection*{References}\label{references}
\addcontentsline{toc}{subsection}{References}

\hypertarget{refs}{}
\hypertarget{ref-ijcci17starcraft}{}
Álvarez-Caballero, Antonio, J. J. Merelo, Pablo García Sánchez, and A.
Fernández-Ares. 2017. ``Early Prediction of the Winner in Starcraft
Matches.'' In \emph{Proceedings Ijcci}.

\hypertarget{ref-buro2012real}{}
Buro, Michael, and David Churchill. 2012. ``Real-Time Strategy Game
Competitions.'' \emph{AI Magazine} 33 (3): 106.

\hypertarget{ref-Stardata17}{}
Lin, Zeming, Jonas Gehring, Vasil Khalidov, and Gabriel Synnaeve. 2017.
``STARDATA: A Starcraft AI Research Dataset.'' In \emph{Proceedings of
the Thirteenth AAAI Conference on Artificial Intelligence and
Interactive Digital Entertainment (Aiide-17), October 5-9, 2017,
Snowbird, Little Cottonwood Canyon, Utah, USA.}, edited by Brian Magerko
and Jonathan P. Rowe, 50--56. AAAI Press.
\url{http://www.aaai.org/Library/AIIDE/aiide17contents.php}.

\hypertarget{ref-ontanon2013survey}{}
Ontanón, Santiago, Gabriel Synnaeve, Alberto Uriarte, Florian Richoux,
David Churchill, and Mike Preuss. 2013. ``A Survey of Real-Time Strategy
Game AI Research and Competition in StarCraft.'' \emph{IEEE Transactions
on Computational Intelligence and AI in Games} 5 (4). IEEE: 293--311.

\hypertarget{ref-Ravari16}{}
Ravari, Yaser Norouzzadeh, Sander Bakkes, and Pieter Spronck. 2016.
``StarCraft Winner Prediction.'' In \emph{Artificial Intelligence in
Adversarial Games: Papers from the Aiide Workshop. Twelfth Artificial
Intelligence and Interactive Digital Entertainment Conference}, 2--8.

\hypertarget{ref-RobertsonW14}{}
Robertson, Glen, and Ian D. Watson. 2014. ``An Improved Dataset and
Extraction Process for Starcraft AI.'' In \emph{Proceedings of the
Twenty-Seventh International Florida Artificial Intelligence Research
Society Conference, FLAIRS 2014, Pensacola Beach, Florida, May 21-23,
2014.}, edited by William Eberle and Chutima Boonthum-Denecke. AAAI
Press.
\url{http://www.aaai.org/ocs/index.php/FLAIRS/FLAIRS14/paper/view/7867}.

\hypertarget{ref-Synnaeve12}{}
Synnaeve, Gabriel, and Pierre Bessière. 2012. ``A Dataset for Starcraft
AI \& an Example of Armies Clustering.'' \emph{CoRR} abs/1211.4552.
\url{http://arxiv.org/abs/1211.4552}.

\hypertarget{ref-Weber09}{}
Weber, Ben George, and Michael Mateas. 2009. ``A Data Mining Approach to
Strategy Prediction.'' In \emph{Proceedings of the 2009 IEEE Symposium
on Computational Intelligence and Games, CIG 2009, Milano, Italy, 7-10
September, 2009}, edited by Pier Luca Lanzi, 140--47. IEEE.
\url{http://ieeexplore.ieee.org/xpl/mostRecentIssue.jsp?punumber=5278415}.

\hypertarget{ref-yoon2001starcraft}{}
Yoon, Sunny. 2001. ``When the Starcraft Launches on the Other Side of
Planet: An Ethnographic Study of the Network Game in Korea.''
\emph{Korea Journalism Studies} 45 (2): 316--437.

\end{document}